\documentclass{INTERSPEECH2023}
\usepackage{CJKutf8} 
\usepackage{multirow} 
\usepackage[utf8]{inputenc} 

\interspeechcameraready

\title{Boosting Chinese ASR Error Correction with Dynamic Error Scaling Mechanism}
\name{Jiaxin Fan$^1$$^,$$^2$, Yong Zhang$^1$, Hanzhang Li$^1$$^,$$^2$, Jianzong Wang$^1$*, Zhitao Li$^1$, Sheng Ouyang$^1$,\\ 
Ning Cheng$^1$, Jing Xiao$^1$\thanks{*Corresponding author: Jianzong Wang, jzwang@188.com.}}

\address{
  $^1$Ping An Technology (Shenzhen) Co., Ltd., China
  \\$^2$Lanzhou University, China
  }

\email{
jzwang@188.com}

\begin{document}

\maketitle
 
\begin{abstract}
Chinese Automatic Speech Recognition (ASR) error correction presents significant challenges due to the Chinese language's unique features, including a large character set and borderless, morpheme-based structure. Current mainstream models often struggle with effectively utilizing word-level features and phonetic information. This paper introduces a novel approach that incorporates a dynamic error scaling mechanism to detect and correct phonetically erroneous text generated by ASR output. This mechanism operates by dynamically fusing word-level features and phonetic information, thereby enriching the model with additional semantic data. Furthermore, our method implements unique error reduction and amplification strategies to address the issues of matching wrong words caused by incorrect characters. Experimental results indicate substantial improvements in ASR error correction, demonstrating the effectiveness of our proposed method and yielding promising results on established datasets.

\end{abstract}

\vspace{-1.2mm}
\section{Introduction}
Automatic speech recognition (ASR) has become a widely used technology for transcribing spoken language into text. However, ASR systems are only sometimes accurate and often make errors in recognizing spoken Chinese. Text error correction techniques have emerged as effective means to address phonetic errors in the text output of ASR models\cite{anantaram2018repairing, mani2020asr, leng2021fastcorrect, fang2022non}. Several challenges make Chinese language error correction particularly difficult. Firstly, the Chinese character set is relatively large, with 3,500 commonly used characters and around 420 pronunciations\cite{duan2019automatically}, requiring more comprehensive error correction methods. Secondly, Chinese lacks clear word boundaries and only uses single words or characters, making it challenging to detect spelling errors and extract contextual semantics. Finally, Chinese words are morpheme-based\cite{bao2020chunk} and typically consist of one to four characters, much shorter than English words. Errors in the phonetic similarity of the semantic center word in Chinese sentences can greatly impact overall semantic accuracy. This poses a challenge for recognizing phonological similarity errors with acoustic models, requiring language models to provide semantic aid.

Recent studies have tackled the problem of error correction in various ways. One popular approach is to treat it as a machine translation problem using a sequence-to-sequence (Seq2Seq)\cite{guo2019spelling, hrinchuk2020correction} framework. In this approach, misspelled sentences generated by an automatic speech recognition (ASR) system are taken as input, and a corrected sentence of the same length is generated as the output. Researchers have also used BERT-based\cite{cui} models for text correction, generating corrected characters for all input characters in parallel.

To improve the performance of contextual spelling correction, efforts have primarily focused on two issues. 
Firstly, researchers have aimed to enhance the quality of language models by introducing external word knowledge. For example, some studies propose incorporating phrase and entity knowledge\cite{zhang2021correcting} for error correction. Additionally, some studies\cite{wu2022spelling} introduced part-of-speech (POS) features and semantic class features to enhance the performance of the model and proposed an auxiliary task to predict the part-of-speech sequence of the target sentence. Other studies\cite{li2022wspeller} have assisted the correction module by predicting the correct word segmentation boundaries from sentences containing misspellings and modifying the input of the embedding layer to incorporate word segmentation information.
Secondly, researchers have explored leveraging phonetic information for text error correction. Some methods consider phonological similarities between pairs of characters, achieved through the increased decoding probability of characters with similar pronunciation or integrating such similarities into the encoding process via graph convolutional networks (GCNs\cite{cheng2020spellgcn}). Another utilization involves directly considering individual character pronunciation, specifically the pinyin. This approach encodes the pinyin of input characters to produce additional phonetic information\cite{xu2021read} or decodes the pinyin of target correct characters to serve as an auxiliary prediction task\cite{liu2021plome,ji2021spellbert}.

However, effective word-level feature fusion remains a challenge due to the risk of miscorrection caused by typos interfering with word segmentation results. Moreover, existing models need to more fully exploit the phonetic information of misspelled characters to ensure accurate predictions.

\begin{figure*}[t]
  \centering
  \includegraphics[width=\linewidth]{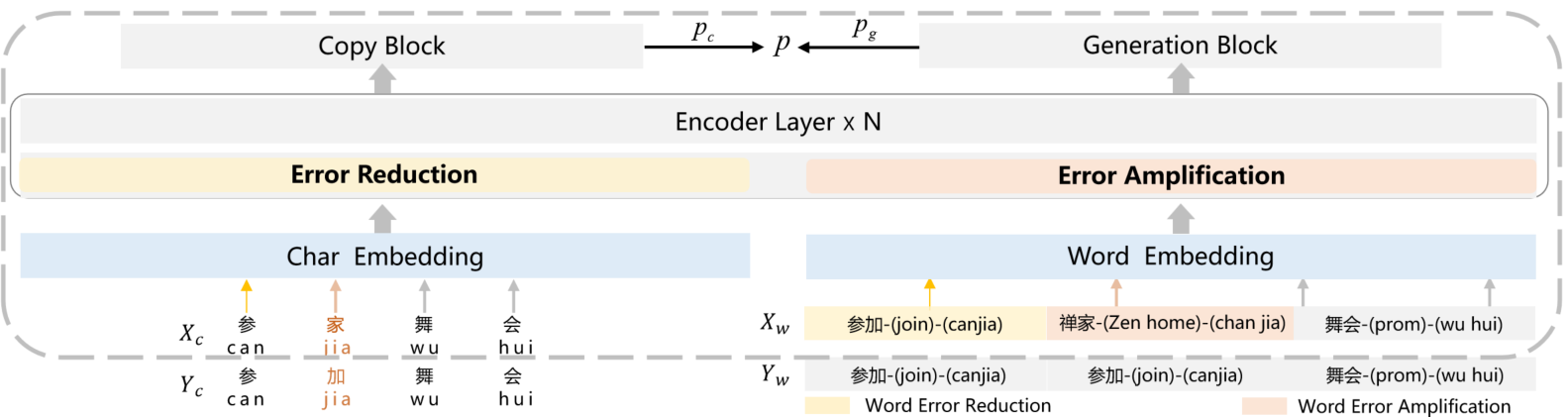}
  \caption{Chinese ASR error correction model based on dynamic error scaling mechanism}
  \label{fig:speech_production}
  \vspace{-4mm}
\end{figure*}

In this paper, we present a novel approach to text error correction based on a dynamic error scaling mechanism that overcomes the limitations of current methods. Our model offers two significant innovations that improve its performance.

Firstly, we explicitly model external lexical knowledge through targeted error reduction and amplification strategies. Unlike previous models that may overcorrect valid tokens or miss errors due to a lack of word-level information, our model prioritizes possibly wrong tokens during encoding and fixes potentially valid tokens during generation to avoid overcorrection.

Secondly, our model leverages word-level pinyin information to identify phonetic candidates instead of relying on character-based matching. However, character-based matching can lead to incorrect candidates, so we introduce a dynamic error scaling mechanism to more accurately identify related words. This approach allows us to explicitly leverage phonetic information and improve the accuracy of our model.

We put forth an efficient and straightforward text error correction approach in this paper. Our model, trained on the extensively used benchmark datasets, SIGHAN2013-2015\cite{wu2013chinese, yu2014overview, tseng2015introduction}, surpasses existing baselines when evaluated on the SIGHAN2015 test set, proving its competitiveness. One of the significant attributes of our approach is its simplicity compared to other sophisticated methods prevalent in the field. The effectiveness and robustness of our model are further exemplified through our comprehensive ablation study.

We highlight three major contributions to the field of text error correction in our work:
\begin{enumerate}
\item We present a unique dynamic error scaling mechanism adept at addressing word-level errors in erroneous texts.
\item We put forward a bi-directional 2-gram pinyin match that accurately harnesses the phonetic information of words, thus helping to identify plausible corrections.
\item Our model's superior performance over the existing baselines on the SIGHAN2015 test set demonstrates the effectiveness and competitiveness of our proposed methodology.
\end{enumerate}
\vspace{-1.2mm}
\section{Method}


In this section, we delineate our proposed method, providing its implementation details and model structure (refer to Figure 1). Our approach comprises a dynamic error scaling mechanism and a correction module, both contributing uniquely to the detection and rectification of errors.

The dynamic error scaling mechanism implements targeted error reduction and amplification strategies, while a bi-directional 2-gram pinyin match extracts the phonetic information of words. Our model processes both incorrect character sentences $X_{c}$ and corresponding word-matching sequences $X_{w}$ as input. After $X_{c}$ is fed into the model, $X_{w}$ is procured via the error scaling module, integrating a char-word attention mechanism to combine character and word-matching sequence information.

The correct versions, $Y_c$ and $Y_w$ represent the targeted correction outcomes and ideal word-matching results, aiding in distinguishing between incorrect and correct word-matching sequences. In the correction module, we generate the correct result, employing a copy distribution simultaneously to avoid over-correction and improve accuracy.

In conclusion, our error correction model operates as follows: Given a character sequence $X_{\mathrm{c}}=$ $\left\{x_{1}^{c}, x_{2}^{c}, \ldots, x_{n}^{c}\right\}$  of length $n$, the model strives to produce a target sequence $Y_{c}=\left\{y_{1}^{c}, y_{2}^{c}, \ldots, y_{n}^{c}\right\}$ of equivalent length, where the typos in $X_{c}$ are rectified. This task is posed as a conditional generation problem, aiming to maximize the conditional probability $P\left(Y_{c} \mid X_{c}\right)$.

\begin{figure}[t]
  \centering
  \includegraphics[width=\linewidth]{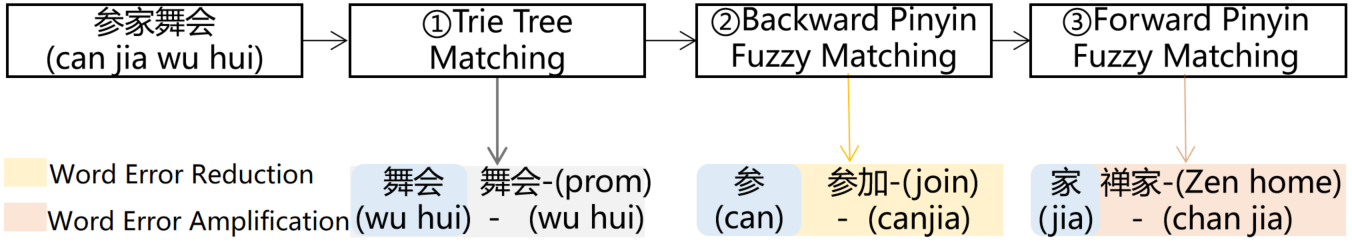}
  \caption{Bi-direction 2-gram pinyin match flow chart of dynamic error scaling mechanism}
  \label{fig:speech_production}
  \vspace{-7mm}
\end{figure}

  \vspace{-1.2mm}

\subsection{Dynamic Error Scaling Mechanism}

This section introduces a dynamic error scaling mechanism aimed at tackling inaccurate word matches resulting from character errors. In particular, the module integrates dictionary information into the model directly, extending character sequences into character-word pairs\cite{liu-etal-2021-lexicon}. For enhancing the matching of incorrect words, we apply both error reduction and amplification strategies and integrate word-level phonetic information via bidirectional pinyin matching.
\vspace{-1.2mm}

\subsubsection{Bi-directional 2-gram Pinyin Matching}
In our method, as illustrated in Figure 2, we utilize the dynamic error scaling mechanism to improve word-matching accuracy. The process begins with trie tree matching (TTM) performed on all characters. For incorrect words resulting from incorrect characters, we incorporate a bidirectional 2-gram Pinyin match for word matching.

To accomplish this, we employ the PyPinyin library to obtain the Pinyin pronunciation of each character. Given that Chinese words commonly consist of 2-gram combinations, our algorithm pairs characters in incorrect words with corresponding 2-gram words from the dictionary in both backward and forward directions. It is important to note that the bidirectional 2-gram Pinyin matching algorithm considers 2-gram word matches in both the input text and the dictionary, rather than solely within the input text.
\vspace{-1.2mm}

\subsubsection{Error Reduction}

In our error reduction strategy, we focus on two distinct scenarios to avoid overcorrection that may arise from integrating incorrect word information. Firstly, for those characters in words that are already correct within the sentence, we strive to accurately match them to the correct words as far as possible. This effort helps to preserve the integrity of the correct elements within the text.

Secondly, for correct characters found within incorrect words, we employ a specific set of strategies. The intention here is to ensure that these correct characters are not inappropriately matched to incorrect words due to the erroneous context they appear in. Notably, this strategy considers pronunciation errors in Chinese, such as flat and warped tongues, and front and back nasal sounds, among others. For instance, we utilize the bi-directional 2-gram pinyin match to guide the character "\begin{CJK}{UTF8}{gbsn}参\end{CJK}-(join)-(can)" found in an incorrect word to its correct word counterpart "\begin{CJK}{UTF8}{gbsn}参加\end{CJK}-(join)-(can jia)". This way, we effectively mitigate the risk of overcorrection, thereby preserving the authenticity of the correct characters even within incorrect words.

\vspace{-1.2mm}
\subsubsection{Error Amplification}

In the case of incorrect characters within erroneous words, it's imperative for the model to assimilate the incorrect word-level information to improve its capability in error detection and correction. For this reason, we adopt an error amplification strategy. During the bidirectional 2-gram pinyin match process, we limit the original characters to words with similar pronunciations and match incorrect words to incorrect characters. For example, the incorrect character "\begin{CJK}{UTF8}{gbsn}家\end{CJK}-(home)-(jia)" can be matched to "\begin{CJK}{UTF8}{gbsn}禅家\end{CJK}-(Zen home)-(chan jia)" for error amplification.

\vspace{-1.2mm}
\subsubsection{Dynamic Char-Word Attention}
After we get the word sequence of all characters, we introduce a character-to-word dynamic attention\cite{vaswani2017attention} mechanism to determine the degree of correlation between each character and its matching. Specifically, for all words assigned to the $i$-th character, we denote their corresponding word vectors as $h_{i}^{w}=\left\{\mathrm{h}_{\mathrm{i} 1}^{\mathrm{w}}, \ldots, \mathrm{h}_{\mathrm{im}}^{\mathrm{w}}\right\}$. Here, $h_{i j}^{w}$ represents the vector of the $j$-th word assigned to the $i$-th character, where $m$ is the total number of words assigned. The relevance of each word ${a}_{i}$ can be calculated by (1), where $\mathrm{W}_{\mathrm{attn}}$ is the weight matrix of bi-linear attention, $h{ }_{i}^{c}$ is the $i$-th  character vector. Through the application of nonlinear transformations to $h_{i}^{w}$, we can align different representations of character and word vectors.
This approach enables our model to capture better and integrate contextual information of word-level features, thereby improving its text error correction ability.
\begin{equation}
{a}_{i}=\operatorname{softmax}\left(h_{i}^{c} \mathbf{W}_{a t t n} \left({W_w}h_{i}^{w}+\mathrm{b_w}\right)^{T}\right)
  \label{eq1}
\end{equation}
Therefore, the weighted sum of all words can be obtained by:
\begin{equation}
  z_{i}^{w}=\sum_{j=1}^{m} a_{i j}\left({W_{j}}h_{ij}^{w}+\mathrm{b_{j}}\right)
  \label{eq2}
    \vspace{-2mm}
\end{equation}

Finally, the weighted dictionary information is injected into the character vector via:
%
\begin{equation}
  \tilde{h}_{i}=h_{i}^{c}+z_{i}^{w}
  \label{eq3}
  \vspace{-2mm}
\end{equation}

\vspace{-3mm}
\subsubsection{Differences From Confusion Sets}
In distinguishing itself from traditional confusion sets, our dynamic error scaling mechanism introduces two innovative features. Firstly, our model uniquely utilizes a 2-gram pinyin match to explicitly leverage the phonetic information, thereby significantly enhancing its performance in correcting ASR output text. Secondly, we employ an innovative strategy of integrating word-level features dynamically. By incorporating a dictionary, our model successfully captures implicit syntactic and semantic knowledge, which markedly bolsters its error detection and correction capabilities.

\vspace{-1.2mm}
\subsection{Correction Module}
Finally, we employ a correction module to correct erroneous sentences, which combines word-level features and syntactic structure to correct errors, effectively improving the error correction accuracy.
The final output of the correction module is a weighted sum of the generative distribution and the copy distribution.

\vspace{-1.2mm}
\subsubsection{Generative Distribution} 
The generative distribution, $\mathcal{P}_{gen} \in \mathrm{R}^{\mathrm{v}}$ is computed by a single-layer feed-forward network with softmax normalization:

\begin{equation}
\mathcal{P}_{\mathrm{gen}}=\operatorname{softmax}\left(\mathrm{W}_{\mathrm{gen}} 
\tilde{h}_{i}+\mathrm{b}_{\mathrm{gen}}\right)
  \label{eq8}
\end{equation}
where $\mathrm{W}_{gen} \in \mathrm{R}^{\mathrm{v} \times 768}$ and $\mathrm{b}_{gen} \in \mathrm{R}^{768}$ are generative parameters, $\mathrm{v}$ is the size of the vocabulary. 

\vspace{-1.2mm}
\subsubsection{Copy Distribution}  
The copy distribution can directly copy the correct characters in the input sequence and only correct the generation distribution of the wrong characters, which is more suitable for the feature that most of the content of the input sequence in the text error correction task is correct, thus avoiding overcorrection. Denote the index of $x_{i}^{c}$ in the
vocabulary as $\mathrm{id}(\mathrm{x}_{\mathrm{i}}^{\mathrm{c}})$, then the copy distribution
of xi, $\mathcal{P}_{\mathrm{copy}} \in\{0,1\}^{{\mathrm{v}}}$, is a one-hot vector satisfying:
\begin{equation}
\mathcal{P}_{\mathrm{copy}}[\mathrm{c}]=\left\{\begin{array}{ll}
0 & \mathrm{c} \neq \mathrm{idx}\left(\mathrm{x}_{\mathrm{i}}^{\mathrm{c}}\right)
\\
1 & \mathrm{c}=\mathrm{idx}\left(\mathrm{x}_{\mathrm{i}}^{\mathrm{c}}\right)
\end{array}\right.
  \label{eq8}
\end{equation}
\vspace{-1.2mm}
\subsubsection{Output Probability}
The final output of the correction module is a weighted sum of the generator distribution and the copy distribution, where the weights are the copy probabilities learned by the model. The copy probability $\omega$ is computed by a layer-normalized two-layer feed-forward network. The final output distribution $\mathcal{P}$ is calculated by:
%
\begin{equation}
\mathcal{P}=\omega \times \mathcal{P}_{\mathrm{copy}}+(1-\omega) \times \mathcal{P}_{\mathrm{gen}}
\end{equation}

Given a training sample $(X, Y)$, the correction loss is defined as:
\begin{equation}
\mathrm{L}=-\sum \log \mathcal{P}\left(\mathrm{Y}_{\mathrm{i}} \mid \mathrm{X}_{\mathrm{i}}\right)
\end{equation}

where $X_i$ is the $i$-th character of the wrong sentence, $Y_i$ is the corrected character of $X_i$, and $ \mathcal{P}$ is the output distribution.
\vspace{-1.2mm}
\section{Experiment}

\subsection{Datasets and Evaluation Metrics}
The training data consisted of 10K manually annotated samples from SIGHAN \cite{wu2013chinese, yu2014overview, tseng2015introduction} and 271K samples from Wang et al.'s work \cite{wang2018hybrid}. We evaluated our proposed model using the SIGHAN2015 \cite{tseng2015introduction} test dataset, which includes 550 positive samples and 550 negative samples. The negative samples represent text without any typos. Table 1 shows the data statistics. We compared the performance of our detection and correction model with several baseline methods \cite{cheng2020spellgcn,liu2021plome,wang2018hybrid,wang2019confusionset} using character-level\cite{liu2022craspell} precision, recall, and F1 scores.
\begin{table}[h]
  \caption{Statistics of datasets}
  \vspace{-3mm}
  \label{tab:1}
\centering
  \setlength{\tabcolsep}{6.75
  mm}
  \begin{tabular}{ccc}
    \toprule
    \textbf{Training Data} & \textbf{Sent} & \textbf{Errors} \\
    \midrule
    SIGHAN2013   & 350      &343                              \\
    SIGHAN2014  &  6,526    &5,122                             \\
    SIGHAN2015  &  3,174    &3,037                             \\
    Wang271k    &  271,329  &381,962
    \\
    \bottomrule
  \end{tabular}

 \centering
 \setlength{\tabcolsep}{7.6mm}
  \begin{tabular}{ccc}
    \toprule   
    \textbf{Testing data}&\textbf{Sent}&\textbf{Errors}\\
    \midrule
    SIGHAN2013  &1,000   &1,224                           \\
    SIGHAN2014  &1,062   &771                             \\
    SIGHAN2015  &1,100   &703                             \\
    \bottomrule
  \end{tabular}
  \vspace{-7mm}

\end{table}

\vspace{-1.2mm}
\subsection{Baselines}
To evaluate the performance of the proposed method, we compare it with the following baseline methods on the SIGHAN2015 test dataset:

\begin{itemize}
\item SoftMask \cite{zhang2020spelling} improves error detection in BERT by introducing a soft mask strategy.
\item SpellGCN \cite{cheng2020spellgcn} combines the GCN network with BERT to model the relationship between characters.
\item cBERT\cite{liu2021plome} has the same architecture as Bert but is pre-trained with knowledge of misspelled words. 
\item PLOME \cite{liu2021plome} proposes a pre-trained model, cBERT\cite{liu2021plome}, and incorporates phonological and visual features based on sequences of phonemes and strokes.
\item CRASpell \cite{liu2022craspell} constructs a noise modeling module based on cBERT, making their model robust against consecutive spelling errors. The model also includes a copy mechanism to handle over-correction.
\end{itemize}

\vspace{-3mm}
\subsection{Experimental Settings}
Based primarily on the configuration by Zhang et al.\cite{zhang2020spelling}, we fix a maximum sentence length of 512, batch size at 32, and implement a learning rate of 5e-5. Our experimental setup incorporates encoders such as BERT\cite{cui} and cBERT\cite{liu2021plome}, while employing the pre-trained word embedding provided by Song et al.\cite{song-etal-2018-directional}. 

\subsection{Main Results}

Table~\ref{tab:2} displays the character-level performance of our cBERT-based model and its baseline methods on the SIGHAN2015 test set. The results demonstrate that our model achieves comparable character correction scores to CRASpell while exhibiting higher recall and F1 scores in character detection.

\begin{table}[h]
\vspace{-3mm}
\caption{The character-level performance on the SIGHAN2015 test set}
\vspace{-2.2mm}

  \label{tab:2}
  \centering
   \setlength{\tabcolsep}{2.1mm}
  \begin{tabular}{ccccccc}
        \toprule
        \multirow{3}{*}{Method} & \multicolumn{3}{c}{Detection-level} & \multicolumn{3}{c}{Correction-level} \\
                                          & P          & R          & F         & P          & R          & F          \\
        \midrule
\textit{SoftMask} & 75.5 & 84.1 & 79.6 & 96.7  & 81.4      & 88.4      \\

\textit{SpellGCN} & 77.7       & 85.6       & 81.4      & 96.9       & 82.9       & 89.4       \\
 \midrule
 \textit{cBERT}   & 83.0       & 87.8       &85.3      & 96.0       & 83.9       & 89.5       \\  

\textit{PLOME}         & \textbf{85.2}       & 86.8       & 86.0      & \textbf{97.2}       & 85.0         & 90.7       \\
\textit{CRASpell}   & 83.5       & 89.2     & 86.3      & 97.1       & \textbf{86.6}       & \textbf{91.5}       \\
\textit{ours (cBERT)}      &\textbf{83.6}        & \textbf{89.8}       & \textbf{86.6}     & 96.2      &86.3        & 91.0  \\
    \bottomrule
\end{tabular}
\vspace{-5mm}

\end{table}

\vspace{-1.2mm}
\subsection{Ablation Study}

We conducted ablation studies to investigate the contribution of each component in our BERT-based model. The components examined were the copy mechanism, raw trie tree matching (TTM), and the dynamic error scaling mechanism (DESM).

Table 3 reveals that the integration of TTM into BERT improved all metrics over raw BERT and BERT with the copy mechanism. Adding a bi-directional 2-gram pinyin match to TTM, forming the DESM, further boosted all scores except correction precision, which decreased compared to TTM. This indicates our model may over-correct errors for better overall performance.


  \vspace{-2.2mm}




\begin{table}[h]
\caption{Ablation Study on SIGHAN2015: character-level}
\vspace{-2.2mm}

  \label{tab:4}
  \centering
  \setlength{\tabcolsep}{1.45mm}
  \begin{tabular}{lcccccc}
        \toprule\multirow{3}{*}{Method} & \multicolumn{3}{c}{Detection-level} & \multicolumn{3}{c}{Correction-level}  \\
                                          & P          & R          & F         & P          & R          & F          \\
        \midrule
\textit{BERT}   &75.8        &85.5        &80.4       &94.7        &80.9        &87.3  \\
\textit{BERT+copy}      &78.1        &85.8       &81.8       &95.7        &82.1        &88.4  
\\
\textit{BERT+TTM+copy}    &80.0        &86.5        &83.1      &\textbf{96.2}   &83.2   &89.2 \\ 

\textit{BERT+DESM+copy}    &\textbf{80.7}        &\textbf{87.6}       &\textbf{84.0}       &95.6  &\textbf{83.8}       &\textbf{89.3}  \\
    \bottomrule
  \end{tabular}
  \smallskip
  \small
\emph{Note: \textit{BERT} and \textit{BERT+copy} data from CRASpell\cite{liu2022craspell}.}

\end{table}

\vspace{-6mm}

\subsection{Error Analysis}
To analyze prediction errors, we conducted a statistical analysis of the Part-Of-Speech of error words in the SIGHAN2015 test set. Table 4 presents the original error distribution and the distribution after correction by our method based on BERT.

\begin{table}[h]
  \caption{Comparison of Part-Of-Speech Distributions between Incorrect and Corrected Text Using Our Method}
  \vspace{-2.2mm}

  \label{tab:1}
  \centering
  \setlength{\tabcolsep}{1.15mm}
  \begin{tabular}{ccccccc}
    \toprule
    \textbf{}&\textbf{verb}&\textbf{noun}&\textbf{adv}&\textbf{pron}&\textbf{aux}&\textbf{others}\\
    \midrule
    \textit{Source} data&32.1\%	&26.7\%	&10.4\%	&8.1\%	&6.3\%	&16.4\% \\
    \textit{BERT} &3.4\%	&2.1\%	&0.2\%	&2.3\%	&1.5\%	&3.2\% \\
    \textit{ours (cBERT)} &1.9\% &1.5\%	&0.3\%	&2.2\%	&1.7\%	&1.4\% \\
    \bottomrule
  \end{tabular}
  \vspace{-3mm}
\end{table}
We use the words matched by the error scaling mechanism and the Jieba library to judge the Part-Of-Speech of the wrong characters. From a statistical analysis of the distribution of Part-Of-Speech tags for wrong words in the incorrect text, it can be seen that nouns and verbs occur more frequently, which mainly consist of 2-gram words. Our model with a dynamic error scaling mechanism can significantly correct 2-gram word errors, thus reducing this type of error significantly.
However, errors associated with single characters, such as auxiliary words (e.g., \begin{CJK}{UTF8}{gbsn}"啊\end{CJK} (ah)" and personal pronouns (e.g., \begin{CJK}{UTF8}{gbsn}"我\end{CJK} (I)" and \begin{CJK}{UTF8}{gbsn}"你\end{CJK} (you)"), often occur at the beginning or end of a sentence, making them challenging to correct. Additionally, our model performs poorly when encountering successive errors at the end of a sentence.

\vspace{-1.2mm}
\section{Conclusions}
In this work, we've introduced a novel dynamic error scaling mechanism aimed at improving Chinese ASR by detecting and correcting phonetically erroneous outputs. This mechanism dynamically combines word-level features and phonetic information, effectively utilizing a bi-directional 2-gram pinyin match to leverage phonetics. Our strategy employs error reduction and amplification to enhance error handling. Benchmark experiments demonstrate our model's improved F1 performance over existing baselines, attesting to the effectiveness and superiority of our approach in Chinese ASR error correction.

\vspace{-1.2mm}
\section{Acknowledgement}
This paper is supported by the Key Research and Development Program of Guangdong Province under grant No.2021B0101400003. The corresponding author is Jianzong Wang from Ping An Technology (Shenzhen) Co., Ltd (jzwang@188.com).


\bibliographystyle{IEEEtran}
\bibliography{mybib}

\end{document}